\documentclass[11pt,a4paper]{article}
\usepackage[hyperref]{acl2021}
\usepackage{times}
\usepackage{latexsym}
\usepackage{lineno}

\usepackage{microtype}

\usepackage{linguex} 

\usepackage{tikz}
\usepackage{pgf}
\usetikzlibrary{shapes,external}
\usetikzlibrary{shapes.multipart}
\usetikzlibrary{calc, positioning, automata}

\usepackage{amsmath}
\usepackage{amssymb}
\usepackage{mathtools}
\usepackage{proof}

\usepackage{xcolor}
\definecolor{first}{RGB}{82,82,82}
\definecolor{enc}{RGB}{253,192,134}
\definecolor{dec}{RGB}{127,201,127}
\definecolor{dec2}{RGB}{140,220,140}
\definecolor{emb}{RGB}{190,174,212}
\definecolor{methods}{RGB}{140,22,22}

\usepackage{subcaption}
\usepackage{graphicx}
\usepackage{floatrow}

\usepackage{tabularx}
\usepackage{booktabs}
\usepackage{multicol}
\usepackage{multirow}

\usepackage{url}

\newcommand{\duck}{\includegraphics[scale=0.021]{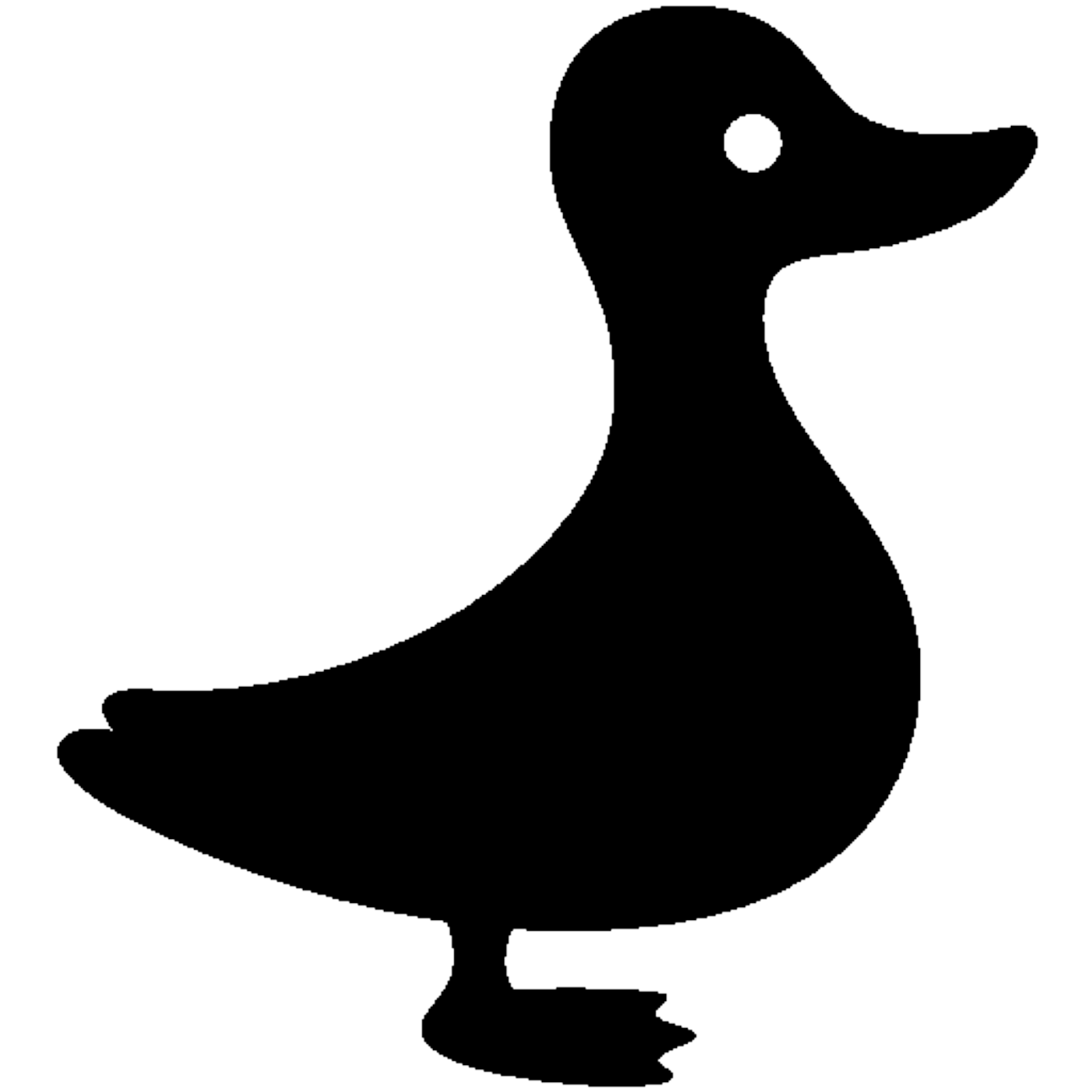}}
\newcommand{\goose}{\includegraphics[scale=0.013]{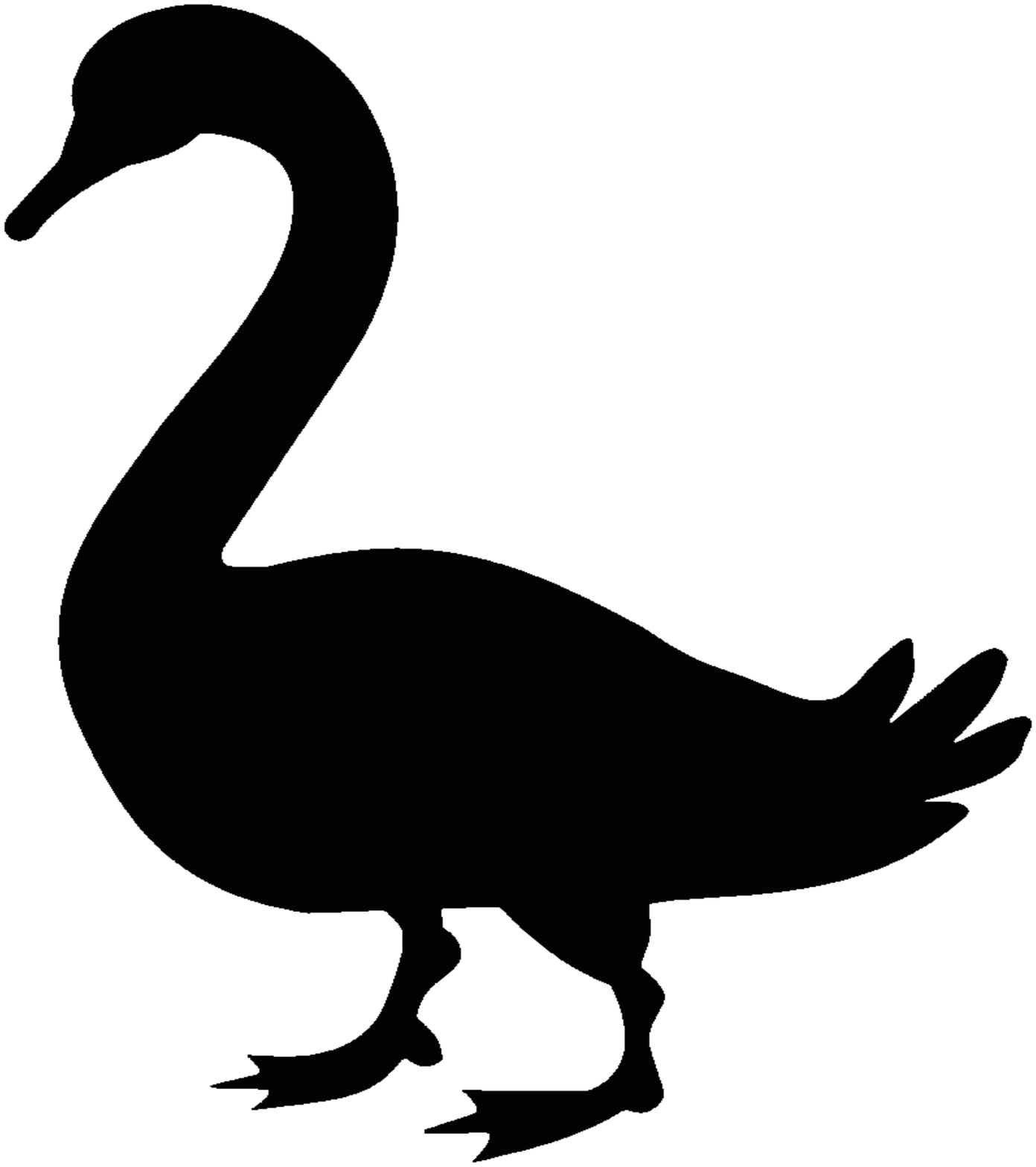}}

\newcolumntype{C}{>{\centering\arraybackslash}X}

\aclfinalcopy 

\title{Improving BERT Pretraining with Syntactic Supervision}

\author{
    Giorgos Tziafas\textsuperscript{{\goose}} \ 
    Konstantinos Kogkalidis\textsuperscript{{\duck}} \ 
    Gijs Wijnholds\textsuperscript{{\duck}} \and
    Michael Moortgat\textsuperscript{{\duck}} \\
    \goose \ University of Groningen\\
    \duck \ Utrecht Institute of Linguistics OTS, Utrecht University \\
    \texttt{g.tziafas@student.rug.nl} \\ 
    \texttt{\{k.kogkalidis,m.j.moortgat,g.j.wijnholds\}@uu.nl}
    }

\begin{document}
\maketitle

\begin{abstract}
Bidirectional masked Transformers have become the core theme in the current NLP landscape.
Despite their impressive benchmarks, a recurring theme in recent research has been to question such models' capacity for syntactic generalization. 
In this work, we seek to address this question by adding a supervised, token-level supertagging objective to standard unsupervised pretraining, enabling the explicit incorporation of syntactic biases into the network's training dynamics.
Our approach is straightforward to implement, induces a marginal computational overhead and is general enough to adapt to a variety of settings.
We apply our methodology on Lassy Large, an automatically annotated corpus of written Dutch.
Our experiments suggest that our syntax-aware model performs on par with established baselines, despite Lassy Large being one order of magnitude smaller than commonly used corpora.
\end{abstract}



\section{Introduction}
In recent years, the advent of Transformers~\cite{vaswani2017attention} has paved the way for high-performing neural language models, with BERT~\cite{devlin2018bert} and its many variants being the main exemplar~\cite{liu2019roberta, sanh2019distilbert,lan2019albert}.
BERT-like models achieve state-of-the-art scores in most major NLP benchmarks via a two-step process.
First, they are trained on massive-scale, minimally processed raw text corpora by employing the so-called masked language modeling (\textit{MLM}) objective.
Task-specific refinements are then obtained by fine-tuning the pretrained model on labeled corpora, usually orders of magnitude smaller in size.

This pipeline, despite its attested performance, suffers from two key limitations.
On the one hand, training a BERT-like model from scratch requires an often prohibitive amount of data and computational resources, barring entry to research projects that lack access to either. 
On the other hand, a naturally emerging question is whether such models develop an internal notion of syntax. 
Discovery of structural biases is hindered by their distributed, opaque representations, requiring manually designed \textit{probing} tasks to extract~\cite{hewitt2019structural,tenney2019you,kim2019pre,clark2019does,goldberg2019assessing,hu-etal-2020-systematic}.
Alternatively, when syntactic evaluation becomes the focal point, it is usually deferred to downstream tasks~\cite{kitaev2019multilingual,ijcai2020-560}, owing both to the lack of sufficiently large labeled corpora as well as the computational bottleneck imposed by hard-to-parallelize operations.

In this work, we seek to alleviate both points by considering them in tandem. 
Contrary to prior work, we consider the case of introducing explicit syntactic supervision during the pretraining process and investigate whether it can allow for a reduction in the data needs of a BERT-like language model.
To facilitate this, we couple the standard unsupervised MLM task with a supervised task, mapping each distinct word to a \emph{supertag}, an abstract syntactic descriptor of its functional role within the context of its surrounding phrase.
In essence, this amounts to simple token-level classification, akin to traditional supertagging~\cite{bangalore1999supertagging}, except for parts of the input now being masked.
In employing both objectives, we ensure that our model is syntax-aware by construction, while incurring only a negligible computational overhead.
We evaluate the trained model's performance in a variety of downstream tasks and find that it performs on par with established models, despite being trained on a significantly smaller corpus.
Our preliminary experiments suggest an improvement to pretraining robustness and offer a promising direction for cheaper and faster training of structure-enhanced language models.
Reflecting on the added objective, we call our model \textit{tagBERT}.

\section{Background}
Embedding structural biases in neural language models has been a key theme in recent research. 
Most syntax-oriented models rely on computationally intensive, hard-to-parallelize operations that constrain their integrability with the state of the art in unsupervised language modeling~\cite{tai-etal-2015-improved,dyer2016recurrent, kim2019unsupervised}.
This can be ameliorated by either asynchronous pretraining, relying on accurate but slow oracles~\cite{kuncoro2019scalable}, or multi-task training, where the system is exposed to a syntactic task for only part of its training routine~\cite{clark2018semi,clark2019bam}.
In the BERT setting, there have been attempts at modifying the architecture by either overlaying syntactic structure directly on the attention layers of the network~\cite{wang-etal-2019-tree} or imposing shallow syntactic cues and/or semantic information in a multi-task setting~\cite{zhang2019semantics,zhou2019limit}. 
While such a setup allows for efficient parallel pretraining, the rudimentary nature of the utilized annotations typically forfeits fine aspects of sentential structure, such as function-argument relations.

In this paper, we adopt lexicalism in the categorial grammar tradition~\cite{ajdukiewicz1935syntaktische,lambek1958mathematics,buszkowski1988categorial,steedman1993categorial,moortgat1997categorial}, according to which (most of) the grammatical structure of a language is encoded in its lexicon via an algebra of types that governs the process of phrasal composition.
Under such a regime, the parse tree underlying a sentence can be partially (or even fully, in the case of an adequately ``strict'' grammar)  recovered from its constituent words and their respective types alone.
In applied terms, the lexical nature of categorial grammars provides us with the opportunity of capturing syntax in a fully-parallel fashion that is straightforward to incorporate with the masked language modeling objective of BERT-like architectures, a fact so far generally overlooked by machine learning practitioners.
This perspective is in line with recent insights arguing for the necessity of explicit supervision for syntactic acquisition~\cite{bailly-gabor-2020-emergence}.

The only prerequisite for our methodology is an adequately sized, categorially annotated corpus.
Even though gold standard corpora exist for a variety of languages and grammars~\cite{chen2004automated,hockenmaier-2006-creating,hockenmaier2007ccgbank,tse2010chinese,ambati2018hindi,kogkalidis2019thel}, their size is generally insufficient for training a parameter-rich neural language model.
This limiting factor can be counteracted by either lexicalizing existing silver-standard corpora of a larger size, or by using an off-the-shelf, high-performance supertagger to annotate the source data prior to pretraining.
In both cases the trained system is likely to inherit common errors of the data-generating teacher; the question is whether the added structural biases facilitate faster training of more general language models, despite potential tagging inaccuracies.

\section{Methodology}
\label{sec:methodology}

\subsection{Data}
To facilitate both the data needs of the neural language model and the added supertagging objective we employ Lassy Large~\cite{vanNoord2013}, a corpus of written Dutch, automatically parsed using the Alpino parser~\cite{bouma2001alpino}. The dataset is comprised of a selection of smaller corpora from varying sources, ranging from excerpts from conventional and modern media to spoken transcripts, enumerating a total of almost 800 million words.
Lassy's syntactic analyses take the form of directed acyclic graphs, with nodes corresponding to words or phrases marked with their part-of-speech as well as syntactic category labels and edges denoting dependency relations.
To make the analyses applicable for our setup, we lexicalize them using the type extraction algorithm of~\citet{kogkalidis2019thel}.
The algorithm traverses a parse graph and encodes its structure in a linear logic proof, under the general paradigm of categorial type logics~\cite{moortgat1997categorial}, simultaneously capturing function-argument and dependency structure.
Words, i.e. fringe nodes in the graph, are assigned \textit{types}, abstract syntactic signs that encode a considerable portion of the full structure.

Applying the extraction algorithm, we obtain a collection of around 66 million sentences, represented as sequences of word-type pairs.
We drop about $20$ million of these in a sanitation step, due to either being duplicates or overlapping with any of the evaluation tasks.
We tokenize words using a preconstructed WordPiece~\cite{wordpiece} vocabulary of $30\,000$ tokens based on a larger collection of written Dutch corpora~\cite{devries2019bertje}.
Further, we keep the $2\,883$ most frequent types, which suffice to cover $95\%$ of the type occurrences in the dataset, and replace the filtered out types with an \texttt{UNK} token.
We finally discard sentences lying in the $5\%$-tail of the length distribution, and train with $45$ million sentences spanning less than $100$ sub-word tokens.

\subsection{Model}
Our model is a faithful replica of BERT\textsubscript{BASE}, except for having a hidden size of $1\,536$ instead of $3\,072$ for the intermediate fully-connected layers, reducing our total parameters from $110$ to $79$ million.
We further employ a linear projection from the model's dimensionality to the number of types in our vocabulary, which we attach to the output of a prespecified encoder block.
The projection can be separably applied on the encoder's intermediate representations, allowing us to optionally query the model for a class weighting over types for each input token.

This addition accounts to a mere $2.5\%$ of the model's total parameter count and only incurs a negligible computational overhead if explicitly enabled, as it does not interfere with the forward pass when the system is run solely as a contextualization model.
If the type classification layer is enabled during pretraining, it introduces a clear error signal that updates all network weights up to the connected encoder block, bolstering the correct acquisition of syntax in the bottom part of the encoding pipeline.

\subsection{Pretraining}
To train our model, we feed it partially masked sentences following the methodology of~\citet{liu2019roberta}; we dynamically mask continuous spans of tokens belonging to the same word and drop the next sentence prediction task, training on single sentences instead.
Attaching the type classification layer at the fourth encoder block, we end up with two output streams.\footnote{The choice of depth for the type classifier is due to preliminary experiments where we let a trainable layer weighter freely select from the range of encoder blocks. In the vast majority of runs, most of the importance was interestingly assigned to the fourth layer.} One is a prediction over the subword vocabulary for each masked token, as in vanilla BERT, whereas
the other comes from the type classifier, yielding a prediction over the type vocabulary for every token, masked or otherwise.%
\footnote{Masking entire words for the supertagging task can be seen as a severe form of regularization, {\`a} la channel dropout.}
We obtain a loss function by summing the cross-entropy between predictions and truths for each output stream.

To deal with the misalignment between subword units and types, we associate every type with the first token of its corresponding word, and mask out predictions spanning subsequent tokens when performing the loss computation.
Similarly, we do not penalize predictions over types discarded by the occurrence count filtering (\texttt{UNK} types).
For regularization purposes, we randomly replace output types $1\%$ of the time~\cite{wu2019stochastic}.

Following standard practices, we optimize using AdamW~\cite{loshchilov2018decoupled} with a batch size of $256$, shuffling and iterating the dataset $8$ times. The learning rate is gradually increased to $10^{-4}$ over $10\,000$ steps and then decayed to zero using a linear warm-up and decay schedule.

\section{Evaluation}
\label{sec:evaluation}
\begin{table*}
    \centering
    \renewcommand{\arraystretch}{1.15}
    \begin{tabular}{@{}l@{\extracolsep{0.7em}}c@{\extracolsep{0.6em}}c@{\extracolsep{1em}}c@{\extracolsep{1em}}c@{\extracolsep{0.7em}}c@{\extracolsep{1em}}c@{\extracolsep{0.5em}}c@{}}
    & \multicolumn{3}{c}{\quad\textit{SoNaR-1}} & \multicolumn{1}{c}{\textit{Lassy UD}} & \multicolumn{1}{c}{\textit{CoNLL}} & \multicolumn{2}{c}{\textit{\AE thel}} \\
    & \small{POS-coarse}
    & \small{POS-fine}
    & \small{NER}
    & \multicolumn{1}{c}{\multirow{1}{*}{\small{POS}}}
    & \multicolumn{1}{c}{\multirow{1}{*}{\small{NER}}}
    & \small{Supertags} & \small{Parse}\\
    \toprule
    \emph{BERTje} \footnotesize{\cite{devries2019bertje}} & $\mathbf{98.8}$ & $\mathbf{97.5}$ & $\mathbf{87.4}$ & $96.4$ & $\mathbf{90.6}$ & $85.5$ & $56.9$\\
    \emph{RobBERT} \footnotesize{\cite{delobelle2020robbert}} & $98.5$ &  $97.2$ & $84.8$ & $96.2$ & $85.9$ & $86.3$ & $56.8$\\
    \emph{tagBERT} \footnotesize{(ours)} & $\mathbf{98.8}$ & $97.4$ & $87.0$ & $\mathbf{96.7}$ & $89.9$ & $\mathbf{86.6}$ & $\mathbf{58.3}$ \\
    \bottomrule
    \end{tabular}
    \caption{Comparative performance for a selection of downstream tasks. We report test set accuracy (\%) on all tasks except NER, where we report F1 scores (\%) as produced by the CoNLL evaluation script~\cite{conll2002}. For a fair comparison, we replicate the fine-tuning process on all pretrained baselines, including truncation of the maximum token length to $100$.}
    \label{table:results}
\end{table*}

To evaluate the trained model, we measure its performance on the below selection of downstream tasks, after fine-tuning.
We keep our fine-tuning set-up as barebones as possible, using Adam~\cite{kingma2014adam} with a batch size of $32$ and a learning rate of $3 \times 10^{-5}$. We apply model selection based on the validation-set performance and report test-set results (averaged over three runs) against the available baselines of each task in Table~\ref{table:results}. In order to provide fair comparisons, we replicate the evaluation of other models using the same experimental setup.

\paragraph{Lassy Small} is a gold-standard syntactically annotated corpus for written Dutch~\cite{vanNoord2013}.
We fine-tune a POS tagger on the subset of the corpus that has been converted to Universal Dependency format~\cite{bouma-van-noord-2017-increasing}.

\paragraph{SoNaR-1}\label{sec:sonar} is a curated subset of Lassy Small that includes several layers of manually added annotations~\cite{sonar1}.
We employ the named entity recognition and part-of-speech labels that come packed with the corpus and treat their classification as downstream tasks. The first contains approximately $60\,000$ samples and $6$ class labels encoded in the IOB scheme, whereas the latter contains about $16\,000$ samples and comes in two varieties: coarse ($12$ classes) and fine-grained ($241$ classes, out of which only $223$ appear in the training data, many just once).

\paragraph{CoNLL-2002} is a named entity recognition dataset from the corresponding shared task~\cite{conll2002}.
The dataset contains $4$ class labels, also encoded in the IOB scheme, with a total size of approximately $24\,000$ samples.

\paragraph{\AE thel} is a typelogical derivation dataset, generated by applying the type extraction algorithm to Lassy Small~\cite{kogkalidis2019thel}.
We replicate the experiments of~\citet{kogkalidis2020neural} to train a typelogical grammar parser, but instantiate the encoder part with the baselines of Table~\ref{table:results}, and report token-level supertagging accuracy as well as full sentential parsing accuracy in the greedy setting.
We note that even though our model is exposed to types during pretraining, their representation format is vastly different during the fine-tuning process; rather than being classification outputs for each word, they are broken down to their primitive symbols and transduced from the input sequence with auto-regressive \textit{seq2seq} decoding. 
In that sense, this task helps us assess the generality of the learned representations.

\section{Discussion}
\label{sec:discussion}
Our model performs on par across all tasks considered, indicating pretraining robustness comparable to the heavy weight baselines of BERT-~\cite{devlin2018bert} and RoBERTa-based~\cite{liu2019roberta} models.\footnote{Implementation code and pretrained model weights will be made available at \url{https://git.io/JOKs4}.}
Considering the non-ideal nature of the silver-standard tags, as well as the significantly smaller size of our corpus compared to competing models, our results can be seen as strong evidence in favor of explicitly encoding structural biases in the pretraining process of neural language models.
Opting for a lexicalized representation of structure allows for a truly seamless and cost-efficient integration with BERT's core architecture, essentially removing the computational bottleneck of alternating between tensor optimization and structure manipulation.

\section{Conclusion}
We introduced tagBERT, a variation of BERT that is biased towards syntax through coupling the standard MLM loss with a supertagging objective.
We trained tagBERT on a modestly sized, silver-standard corpus of written Dutch -- after first lexicalizing its annotations --
and evaluated the trained model on a number of downstream NLP tasks after fine-tuning. 
Despite the corpus' modest size, our method is achieving performance comparable to established state-of-the art models.
This result is contrary to the ongoing trend of utilizing increasingly more data and augmenting model capacity, instead suggesting potential benefits from incorporating richer annotations in convenient representation formats.
Our work aims towards a syntactically-transparent, cost-efficient language model that combines both the rigor of formal linguistic theories and the representational power of large-scale unsupervised learning.

We leave several directions open for future work, including more extensive experimentation with different languages and grammar formalisms, integration with existing pre-trained models in an intermediate-training fashion~\cite{wang-etal-2019-tell} and exploring architectural adjustments that would allow a two-way dependence or a stronger interfacing between the lexical and syntactic modalities, moving towards structurally-conditioned language generation and structure-aware sentence embeddings, akin to~\citet{zanzotto-etal-2020-kermit}.

\bibliography{main,anthology}
\bibliographystyle{acl_natbib}

\end{document}